\documentclass[11pt]{article}

\usepackage[final]{acl}

\usepackage{microtype}
\usepackage{enumitem}

\usepackage{booktabs}
\usepackage{amsmath} 
\usepackage{tikz}
\usetikzlibrary{positioning,arrows.meta}
\usepackage[english,bidi=default]{babel} 
\babelfont{rm}{TeXGyreTermesX} 
\newfontfamily\devanagarifont{NotoSansDevanagari-Regular.ttf}

\title{English-to-Prakrit Machine Translation via Multilingual Transfer Learning}

\author{Om Choksi \and Smit Kareliya \and Shrikant Malviya \and Pruthwik Mishra\\
        Sardar Vallabhbhai National Institute of Technology, Surat\\
        \texttt{\{p25ds028,p25ds025,shrikant\}@coed.svnit.ac.in}
        , \texttt{pruthwikmishra@aid.svnit.ac.in}}

\begin{document}

\maketitle

\begin{abstract}
We study English-to-Prakrit machine translation in a low-resource setting where the target language is unsupported by IndicTrans2. We adapt the multilingual model by mapping Prakrit to the Hindi language tag (\texttt{hin\_Deva}) without modifying the tokenizer, vocabulary, or architecture. Using a 1,474-pair Maharashtri Prakrit parallel corpus and evaluation on a 20-sample Ardhamagadhi test set, we report corpus BLEU improvements over an untuned baseline. The results indicate that script-compatible language routing can enable feasible transfer to unsupported classical languages, while highlighting limitations due to data scarcity and dialect mismatch. Our code and trained models are released to the public for
further exploration 
\href{https://github.com/D3v1s0m/indictrans2-prakrit-mt}{\textcolor{blue}{https://github.com/D3v1s0m/indictrans2-prakrit-mt}}.
\end{abstract}

\section{Introduction}
Prakrit is a family of Middle Indo-Aryan languages that played a central role in classical Indian literature, Jain canonical texts, Buddhist traditions, inscriptions, and dramatic works \cite{pischel1981,woolner1986,ollett2017}. Important historical resources such as Ashokan inscriptions and Jain Āgamas were composed in various Prakrit dialects, making the language family a significant component of South Asia's cultural and intellectual heritage. Despite its cultural importance, machine translation (MT) resources and systems for Prakrit remain scarce \cite{prakhritnlp}. 

Recent multilingual neural machine translation models such as IndicTrans2 have demonstrated strong transfer learning capabilities across a wide range of modern Indic languages \cite{gala2023indictrans2highqualityaccessiblemachine}. This work explores low-resource adaptation of IndicTrans2, a multilingual Transformer MT system for Indic languages, to English-to-Prakrit translation. IndicTrans2 does not natively support Prakrit; therefore, we treat Prakrit as an unsupported target and route it through the Hindi target tag (\texttt{hin\_Deva}), leveraging shared Devanagari script.

In this work, we investigate the feasibility of adapting IndicTrans2 for English-to-Prakrit machine translation in a low-resource setting. We focus on a constrained setting with a small parallel corpus and a cross-dialect evaluation set. The goal is not to claim high accuracy, but to assess feasibility of transfer to unsupported languages using only tag-level adaptation and standard training infrastructure.

Our contributions are:
\begin{itemize}[noitemsep, topsep=0pt]
	\item Adapted IndicTrans2 for unsupported English-to-Prakrit translation using script-compatible tag routing.
	\item Achieved substantial BLEU improvement over the untuned baseline.
	\item Demonstrated cross-dialect transfer from Maharashtri training data to Ardhamagadhi evaluation data.
	\item We release code, models, and experimental pipelines to facilitate future research on Ancient Indic Language Processing.
\end{itemize}

\section{Related Work}
Low-resource MT remains challenging due to limited parallel data and domain mismatch \citep{koehn-knowles-2017-six}. Recent multilingual neural machine translation systems have demonstrated strong transfer learning capabilities through shared model parameters and multilingual representations \citep{johnson-etal-2017-googles}. Large-scale multilingual models such as mBART \cite{liu2020mbart}, NLLB \cite{costa2022nllb}, and IndicTrans2 \cite{gala2023indictrans2highqualityaccessiblemachine} have shown that transfer from high-resource languages can substantially improve translation quality for low-resource languages.

For classical and historical languages, prior work is sparse, often constrained to small datasets and limited evaluation \citep{prakhritnlp}. Available Prakrit datasets \cite{viitpune-dataset,prakhritnlp} are typically small and fragmented, limiting the development and evaluation of modern machine translation systems.

Our work differs from prior research by investigating the adaptation of a state-of-the-art multilingual translation model to an unsupported classical language. Specifically, we explore whether multilingual representations learned by IndicTrans2 can be transferred to Prakrit through script-compatible language routing, without modifying the tokenizer, vocabulary, or model architecture.

\begin{table}[t]
\centering
\begin{tabular}{lcc}
\hline
Split & Samples & Prakrit Variety \\
\hline
Train & 1,326 & Maharashtri \\
Validation & 148 & Maharashtri \\
Test & 20 & Ardhamagadhi \\
\hline
Total & 1,494 & Mixed Dialects \\
\hline
\end{tabular}
\caption{Dataset statistics. Training and validation use Maharashtri Prakrit, while testing is performed on Ardhamagadhi Prakrit, resulting in a cross-dialect evaluation setting.}
\label{tab:data_stats}
\end{table}

\section{Dataset Description}
We use the VIITPune Prakrit-to-English Parallel Corpus \cite{viitpune-dataset}, which contains 1,474 parallel English--Prakrit sentence pairs. The Prakrit side of the corpus is primarily written in Maharashtri Prakrit, one of the major literary varieties of the Prakrit language family. The corpus consists of aligned sentence-level translations and serves as the primary resource for fine-tuning due to the scarcity of publicly available Prakrit parallel datasets.

The dataset was randomly divided into training and validation sets using a 90:10 split, resulting in 1,326 training instances and 148 validation instances. To assess the generalization capability of the model beyond the training dialect, we perform evaluation on a separate Ardhamagadhi Prakrit test set containing 20 sentence pairs obtained from the PrakritNLP dataset \cite{prakhritnlp}. This setup introduces a challenging cross-dialect evaluation scenario, where the model is trained on Maharashtri Prakrit but tested on Ardhamagadhi Prakrit.

Table~\ref{tab:data_stats} summarizes the dataset statistics. The cross-dialect setting provides a preliminary assessment of the robustness of multilingual transfer learning for unsupported Prakrit varieties and reflects realistic conditions where resources from multiple historical dialects must be leveraged jointly.

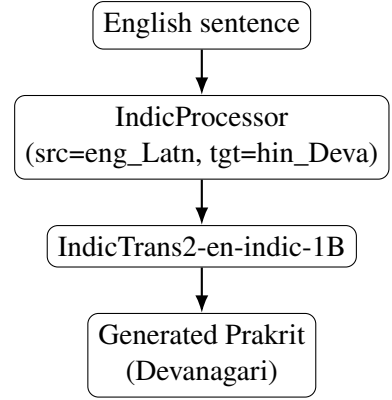
\begin{figure}[t]
\centering
\begin{tikzpicture}[
	node distance=6mm,
	box/.style={draw, rounded corners, align=center, inner sep=4pt},
	arr/.style={-Latex, thick}
]
\node (src) [box] {English sentence};
\node (proc) [box, below=of src] {IndicProcessor\\(src=eng\_Latn, tgt=hin\_Deva)};
\node (model) [box, below=of proc] {IndicTrans2-en-indic-1B};
\node (tgt) [box, below=of model] {Generated Prakrit\\(Devanagari)};
\draw[arr] (src) -- (proc);
\draw[arr] (proc) -- (model);
\draw[arr] (model) -- (tgt);
\end{tikzpicture}
\caption{Compact pipeline for unsupported Prakrit adaptation via Hindi tag routing.}
\label{fig:pipeline}
\end{figure}
\section{Methodology}

Our approach investigates whether a multilingual machine translation model can be adapted to an unsupported classical language through transfer learning without modifying the underlying model architecture. Figure~\ref{fig:pipeline} illustrates the overall translation pipeline.

\subsection{Base Model}
We use \href{https://huggingface.co/ai4bharat/indictrans2-en-indic-1B}
{\texttt{ai4bharat/indictrans2-en-indic-1B}} as the base model. IndicTrans2 is a multilingual Transformer MT system trained on large-scale English--Indic parallel corpora covering 22 scheduled Indian languages.  \citep{gala2023indictrans2highqualityaccessiblemachine}. The model employs language tags to control target language generation and has demonstrated strong performance across a wide range of low-resource and multilingual translation tasks. IndicTrans2 does not provide explicit support for Prakrit. However, its multilingual architecture and shared representation space make it suitable for investigating unsupported Indo-Aryan languages.

\begin{table*}[t]
\centering
\small
\begin{tabular}{p{0.34\linewidth}p{0.28\linewidth}p{0.28\linewidth}}
\toprule
English & Reference Prakrit & Generated Prakrit \\
\midrule
Being separated from what is dear, and being forced to look at what is not, are two causes of grief. I bow to your good breeding which demands that you act as you do.
&
{\devanagarifont किं किं दे पडिहासइ सहीहि इअ पुच्छिआइ मुद्धाए, पढमुग्गअदोहलिणीअ णवर दइअं गआ दिट्ठी}
&
{\devanagarifont किं किं दे पडिहासइ सहीहि इअ पुच्छिआइ मुद्धाए, पढमुग्गअदिण्णं को वि पिअमे दिट्ठीए} \\
\bottomrule
\end{tabular}
\caption{Qualitative English-to-Prakrit translation example generated by the fine-tuned IndicTrans2 model.}
\label{tab:qual}
\end{table*}

\begin{table}[t]
\centering
\small
\begin{tabular}{lr}
\toprule
Model & BLEU \\
\midrule
Baseline IndicTrans2-1B & 1.57 \\
Fine-tuned IndicTrans2-1B & 14.3 \\
\bottomrule
\end{tabular}
\caption{BLEU score on the Ardhamagadhi test set.}
\label{tab:results}
\end{table}

\begin{table}[t]
\centering
\small
\begin{tabular}{lr}
\toprule
Hyperparameter & Value \\
\midrule
Epochs & 20 \\
Learning rate & 1e-5 \\
Batch size & 4 \\
Weight decay & 0.01 \\
Label smoothing & 0.1 \\
Optimizer & AdamW (default) \\
Trainer & Seq2SeqTrainer \\
Training time & 54m 36s \\
\bottomrule
\end{tabular}
\caption{Training hyperparameters.}
\label{tab:hparams}
\end{table}

\subsection{Unsupported Language Adaptation}
Prakrit is not explicitly supported by IndicTrans2. We map Prakrit to the Hindi target tag (\texttt{hin\_Deva}) based on the shared Devanagari script. We do not modify the tokenizer, vocabulary, or architecture. Since IndicTrans2 associates target language tags with multilingual, script-aware representations, the shared script enables usable transfer despite the absence of explicit Prakrit support.

The motivation for selecting Hindi is twofold. First, both Hindi and the available Prakrit corpora are represented in the Devanagari script, allowing direct reuse of the existing tokenizer and vocabulary. Second, Hindi and Prakrit belong to the Indo-Aryan language family and therefore exhibit partial lexical, morphological, and syntactic similarities. We hypothesize that the multilingual representations learned by IndicTrans2 can be transferred to Prakrit through this shared linguistic space.

\subsection{Training Procedure}
We perform full fine-tuning using Hugging Face Transformers \citep{wolf-etal-2020-transformers}. Source and target sentences are preprocessed using \texttt{IndicProcessor}, which handles language-specific normalization and formatting required by IndicTrans2.

For each training instance, the English source sentence is tagged with the source language identifier \texttt{eng\_Latn}, while the target Prakrit sentence is associated with the Hindi target language tag \texttt{hin\_Deva}. The original IndicTrans2 tokenizer and vocabulary are retained without modification.

Model parameters are optimized using the AdamW optimizer \cite{loshchilov2018decoupled} with teacher forcing under a sequence-to-sequence learning objective. Fine-tuning is performed on the Maharashtri Prakrit training corpus, while validation is conducted on a held-out Maharashtri split. Final evaluation is carried out on an Ardhamagadhi Prakrit test set to assess cross-dialect generalization.

The resulting fine-tuned model has been publicly released on Hugging Face\footnote{\href{https://huggingface.co/D3v1s0m/indictrans2-en-prakrit-1B}{\texttt{https://huggingface.co/D3v1s0m/indictrans2-en-prakrit-1B}}} to facilitate reproducibility and future research on Prakrit machine translation.

\subsection{Inference Pipeline}

During inference, an English sentence is first processed by \texttt{IndicProcessor} and tagged as \texttt{eng\_Latn}. The fine-tuned IndicTrans2 model then generates output using the \texttt{hin\_Deva} target tag. The generated text is interpreted as Prakrit and evaluated against the reference translations using corpus-level BLEU scores. This pipeline enables English-to-Prakrit translation while preserving the original IndicTrans2 architecture and multilingual training framework.

\section{Experimental Setup}
Training uses a single NVIDIA A100 80GB GPU. Hyperparameters are shown in Table~\ref{tab:hparams}. The training time is approximately 54m36s. We evaluate using SacreBLEU \citep{post-2018-call,papineni-etal-2002-bleu} with \texttt{sacrebleu.corpus\_bleu(hyps, refs)}.

\section{Results and Evaluation}
Table~\ref{tab:results} reports the corpus-level BLEU scores on the Ardhamagadhi Prakrit test set. Fine-tuning IndicTrans2 on the Maharashtri Prakrit corpus improves BLEU from 1.57 to 14.30, representing a substantial gain over the untuned baseline. Given that Prakrit is not natively supported by IndicTrans2 and that only 1,326 training sentence pairs are available, this improvement demonstrates the effectiveness of multilingual transfer learning for extremely low-resource classical languages.

The observed improvement suggests that the multilingual representations learned by IndicTrans2 can be transferred to unsupported Indo-Aryan languages through script-compatible language routing. Although the model was originally trained for modern Indic languages, the shared Devanagari script and linguistic similarities between Hindi and Prakrit appear sufficient to facilitate meaningful adaptation without modifying the tokenizer, vocabulary, or model architecture.

We also tested a smaller 200M model; it performed poorly and was not used further. The cross-dialect setting likely contributes to the lower absolute BLEU and indicates the need for dialect-aware adaptation strategies.

A qualitative translation example is shown in Table~\ref{tab:qual}. The generated output exhibits several lexical and structural similarities to the reference translation, indicating that the model is capable of producing plausible Prakrit sequences. However, noticeable differences remain in vocabulary selection and grammatical realization, reflecting the challenges posed by limited training data and dialectal variation between Maharashtri and Ardhamagadhi Prakrit.

\section{Conclusion}
This paper investigated the feasibility of adapting IndicTrans2 for English-to-Prakrit machine translation in an unsupported low-resource setting. By reusing the \texttt{hin\_Deva} language tag without modifying the tokenizer, vocabulary, or model architecture, we demonstrated that multilingual transfer learning can produce meaningful improvements in translation quality, even with limited training data and a challenging cross-dialect evaluation setup. The results suggest that script-compatible adaptation offers a simple yet effective approach for extending modern multilingual translation models to historical Indo-Aryan languages. Although translation quality remains constrained by data scarcity and dialectal variation, the findings provide encouraging evidence for the development of AI-based tools for Prakrit and other ancient languages, representing an initial step toward broader Ancient Indian Language Processing research.

\section{Limitations}
The study is limited by the small size of the available Prakrit parallel corpus and the very small evaluation set. We do not perform human evaluation, linguistic validation by Prakrit experts, or detailed analysis of Hindi lexical leakage in the generated translations. The dialect mismatch between Maharashtri training data and Ardhamagadhi evaluation data may further confound performance estimates. In addition, we evaluate only a single adaptation strategy based on the \texttt{hin\_Deva} language tag and do not compare alternative routing approaches using other Indic language tags. Finally, the work does not explore data augmentation, domain adaptation, or dedicated Prakrit language modeling, all of which may further improve translation quality.


\bibliography{references}

\end{document}